\documentclass[letterpaper, 10 pt, conference]{ieeeconf}  % Comment this line out if you need a4paper

\IEEEoverridecommandlockouts                              % This command is only needed if 
\usepackage[dvipsnames]{xcolor}  % Load xcolor first, with options
\usepackage{graphicx}            % Only once
\usepackage{tikz}
\usepackage{etoolbox}
\usepackage{booktabs}
\usepackage{multirow}
\usepackage{amsmath}
\usepackage{verbatim}
\usepackage{colortbl}
\usepackage{amsfonts}
\usepackage{csquotes}
\usepackage{hyperref}
\usepackage{subfig}
\usepackage{placeins}
\usepackage{url}

\title{\LARGE 
\bf
 MambaAct: Efficient Mamba Action Experts for Flow-Matching Vision-Language-Action Models}
 \author{Khuyen Pham$^{1}$, Tran Nguyen Le$^{2}$, Edoardo Fazzari$^{1*}$, Cesare Stefanini$^{1}$
 \thanks{
 $^{*}$ Corresponding author: edoardo.fazzari@mbzuai.ac.ae;
 $^{1}$ MBZUAI;
 $^{2}$ Technical University of Denmark.}
 }

\begin{document}

\maketitle
\thispagestyle{empty}
\pagestyle{empty}
%%%%%%%%%%%%%%%%%%%%%%%%%%%%%%%%%%%%%%%%%%%%%%%%%%%%%%%%%%%%%%%%%%%%%%%%%%%%%%%%

% create your cmd to include comments and discuss in the text 

\newcommand{\ppar}[1]{{\color{magenta} [ppar]: #1}}

\newcommand{\plund}[1]{{\color{cyan} [plund]: #1}}

\newcommand{\dtai}[1]{{\color{yellowgreenn} [dtai]: #1}}

% \begin{figure*}[!t]
%     \centering
%     \includegraphics[width=0.9\textwidth]{figures/MambaAct.pdf}
%     \caption{
%       \textbf{MambaAct architecture.}
%       MambaAct retains the $\pi_{0.5}$ visual-language backbone (later jointly fine-tuned with the new Mamba action expert), including the SigLIP vision encoder and Gemma prefix layers, and replaces the original
%       transformer action expert with a compact Mamba action expert.
%       The VLM prefix tokens are compressed into context tokens $C$, composed of vision, language, and learned-query global summary tokens.
%       A masked mean pooling operation maps $C$ to a condition vector $z$.
%       For each Mamba block, $z$ is passed through a FiLM gate MLP to generate scale, shift, and gate parameters that modulate the Mamba residual update.
%       During flow-matching denoising, the Mamba action expert processes the noisy action chunk, conditioned on flow time, to predict a flow velocity that is
%       iteratively denoised into the continuous action chunk.
%     }
%     \label{fig:mambaact-architecture}
% \end{figure*}

\begin{abstract}
% Vision-language-action (VLA) models have shown strong promise for generalist robotic manipulation, but their inference cost remains a major bottleneck for real-time control. This limitation is especially pronounced in flow-matching VLAs such as $\pi_{0.5}$, where the action expert is evaluated repeatedly during iterative denoising. Existing efficiency approaches primarily reduce sampling steps, compress or prune visual-language backbones, or redesign the overall policy into smaller architectures, leaving the repeatedly evaluated action expert comparatively underexplored. We introduce MambaAct, an efficient Mamba action expert for flow-matching VLA policies. MambaAct retains the pretrained visual-language prefix encoder of $\pi_{0.5}$, while replacing its transformer-based action expert with a compact Mamba-based action expert. Within this expert, Mamba serves as the action mixer that models dependencies across the noisy continuous-action sequence. To preserve visual-language conditioning without repeated token-level cross-attention, MambaAct uses a lightweight Context Compressor to map Vision-language model (VLM) prefix tokens into a compact conditioning signal that gates and modulates each Mamba action block. MambaAct uses a $10\times$ smaller action expert and reduces flow-matching denoising latency by $64.8\%$ on LIBERO and $58.8\%$ in real-world deployment, while preserving or improving policy performance; across three LIBERO training seeds, it improves average success by 1.08 percentage points over the matched $\pi_{0.5}$ baseline.
Vision-language-action (VLA) models have shown strong promise for generalist robotic manipulation, but their inference cost remains a major bottleneck for real-time control. This limitation is especially pronounced in flow-matching VLAs such as $\pi_{0.5}$, where the action expert is evaluated repeatedly during iterative denoising. We introduce MambaAct, an efficient Mamba action expert for flow-matching VLA policies. MambaAct retains the pretrained visual-language prefix encoder of $\pi_{0.5}$, while replacing its transformer-based action expert with a compact Mamba-based action expert. Within this expert, Mamba serves as the action mixer that models dependencies across the noisy continuous-action sequence. To preserve visual-language conditioning without repeated token-level cross-attention, MambaAct uses a lightweight Context Compressor to map Vision-language model (VLM) prefix tokens into a compact conditioning signal that gates and modulates each Mamba action block. MambaAct uses a $10\times$ smaller action expert and reduces flow-matching denoising latency by $64.8\%$ on LIBERO and $58.8\%$ in real-world deployment, while improving mean LIBERO success by $1.08$ percentage points across three independent fine-tuning seeds and maintaining gains on LIBERO-Plus and real-world benchmarks. Our code and model weights are publicly available at \url{https://github.com/giakhuyendihoc/MambaAct}.
% Across LIBERO and real-world UR5 manipulation tasks, MambaAct improves policy success over matched $\pi_{0.5}$ baselines while using a $10\times$ smaller action expert and reducing flow-matching denoising latency by $64.8\%$ on LIBERO and $58.8\%$ on the real-world tasks.

\end{abstract}

\section{Introduction}
Recent flow-matching VLAs such as $\pi_0$~\cite{black2024pi_0} and $\pi_{0.5}$~\cite{black2025pi_0} achieve strong manipulation performance, but their iterative action-generation process introduces substantial inference cost. At every policy query, the visual-language prefix can be computed once, whereas the action expert must be evaluated repeatedly across multiple flow-matching denoising steps. As a result, action generation can become the dominant source of latency even when the visual-language backbone itself is relatively compact.

Recent efforts to improve VLA efficiency have largely reduced this cost by decreasing the number of sampling steps, compressing or pruning pretrained networks, reducing visual-token processing, or redesigning the policy around smaller architectures. While effective, these approaches primarily accelerate inference by reducing how often existing components are evaluated or by simplifying the visual-language pathway and overall model. In contrast, the architecture of the repeatedly evaluated action expert itself remains comparatively underexplored as an efficiency axis. This motivates us to ask whether flow-matching inference can instead be accelerated by replacing the action-generation module while preserving the pretrained visual-language representations and standard flow-matching formulation.

We address this question with \textbf{MambaAct}, a compact Mamba~\cite{lahotimamba} action expert for flow-matching VLA policies. MambaAct retains the pretrained $\pi_{0.5}$ visual-language encoder and replaces its large Transformer-based action expert with a lightweight Mamba sequence mixer. Mamba is attractive in this setting because its selective state-space formulation provides a compact inductive bias for modeling ordered, strongly correlated action sequences. As we show empirically, this design enables aggressive action-expert compression while preserving policy performance. To efficiently condition the newly initialized expert on the pretrained VLM, we introduce a Context Compressor and gated FiLM~\cite{perez2018film} interface that transform the visual-language prefix into a compact representation reused throughout flow-matching denoising. The resulting expert contains only $29.7$M parameters, compared with $311.5$M for the original $\pi_{0.5}$ action expert.

Our main contributions are:
 \begin{itemize}
 \item We introduce \textbf{MambaAct}, establishing action-expert replacement as a complementary efficiency axis for flow-matching VLAs. MambaAct replaces the repeatedly evaluated Transformer action expert with a compact Mamba expert while retaining the pretrained visual-language backbone and standard flow-matching formulation, without additional pretraining or distillation.
% \item We demonstrate a $\sim$$10\times$  reduction in action-expert parameters and substantial denoising acceleration while matching or improving policy performance. 

% \item MambaAct uses over $10\times$ fewer action-expert parameters than $\pi_{0.5}$ while matching or improving policy performance.
% \item We show that Mamba enables aggressive action-expert compression: MambaAct uses over $10\times$ fewer action-expert parameters than $\pi_{0.5}$  while matching or improving policy performance.
\item We show that Mamba enables aggressive action-expert compression: MambaAct uses over $10\times$ fewer action-expert parameters than $\pi_{0.5}$ while improving mean LIBERO~\cite{liu2023libero} success by $1.08$ percentage points across three fine-tuning seeds.
% Across LIBERO~\cite{liu2023libero}, LIBERO-Plus~\cite{fei2025libero}, and real-world UR5 manipulation tasks, it reduces flow-matching denoising latency by $64.8\%$ in simulation and $58.8\%$ in real-world deployment.
 
\item Through controlled ablations, we examine the contributions of the Mamba action mixer and conditioning interface to the efficiency-performance trade-off. We further demonstrate that MambaAct composes with VLM-depth reduction and transfers to SmolVLA~\cite{shukor2025smolvla}, indicating that action-expert replacement generalizes across flow-matching VLA backbones.

\end{itemize}

\section{Related Work}
\subsection{Efficient Vision-Language-Action Models}
The growing computational cost of VLA models has motivated a range of approaches that reduce inference cost at different stages of the policy. One direction focuses on accelerating iterative action generation. SnapFlow~\cite{luan2026snapflow}, for example, uses progressive self-distillation to compress iterative flow matching into a single action-expert evaluation, thereby reducing the number of sampling steps required at inference time. Complementary approaches instead reduce the cost of each network evaluation by simplifying the underlying architecture. Shallow-$\pi$~\cite{jeon2026shallow} distills shallower visual-language and action transformers, while CLP~\cite{nguyen2026finetuning} identifies and removes redundant layers from pretrained VLA models.

Another line of work targets the visual-language pathway, where long visual-token sequences can dominate computation. General VLM pruning methods such as FastV~\cite{chen2024image} and DivPrune~\cite{alvar2025divprune} preserve early visual processing while removing redundant tokens before later transformer layers, and have subsequently been evaluated in VLA policies~\cite{nguyen2026finetuning}. VLA-specific methods further adapt this idea to robotic control: LightVLA~\cite{jiang2025better}, ADP~\cite{pei2025action}, and SpecPrune-VLA~\cite{wang2025specprune} selectively reduce visual tokens while retaining task-relevant information for action prediction. Beyond pruning individual components, several works pursue efficiency through compact architectures or coordinated reductions across multiple computational axes. SmolVLA~\cite{shukor2025smolvla} reduces the overall model scale, whereas CogVLA~\cite{li2026cogvla} combines instruction-conditioned visual aggregation, token sparsification, and a redesigned action decoder. Similarly, EfficientVLA~\cite{yang2026efficientvla} combines token and layer reduction with feature reuse across action-generation steps.

MambaAct targets a different design point: it retains the pretrained $\pi_{0.5}$ backbone and standard flow-matching formulation, but replaces the transformer action expert with a compact Mamba-based expert trained jointly during task fine-tuning, without teacher models, distillation objectives, or additional pretraining.

\subsection{Mamba in Vision-Language-Action Models}
Mamba~\cite{lahotimamba} replaces pairwise self-attention with selective state-space recurrence, providing a compact mechanism for modeling ordered sequences. This makes it attractive for repeatedly evaluated action experts operating under tight parameter budgets. Mamba has been explored across sequence modeling and computer vision~\cite{FAZZARI2025102955}, with recent extensions to robotic perception and control~\cite{liu2024robomamba,cao2025mamba,islam2026sureflow}. 
Within VLA modeling, RoboMamba~\cite{liu2024robomamba} uses Mamba primarily as the multimodal visual-language reasoning backbone, targeting efficient perception and reasoning rather than dense continuous action generation. 

Mamba has also been explored for continuous control. Mamba Policy~\cite{cao2025mamba} applies Mamba to continuous-action diffusion, demonstrating selective state-space models as compact action-denoising networks. However, it is not designed as an action expert conditioned on representations from a pretrained vision-language backbone. More recently, SUREFlow~\cite{islam2026sureflow} develops a lightweight Mamba-based multimodal policy with uncertainty-aware residual flow matching, jointly predicting action velocity and uncertainty to selectively refine unreliable actions. Unlike MambaAct, SUREFlow constructs a compact end-to-end multimodal policy rather than preserving a pretrained foundation VLM and replacing only its action expert.

MambaAct targets this complementary setting: accelerating the action-generation pathway of a pretrained flow-matching VLA while preserving its learned visual-language representations. It retains the $\pi_{0.5}$ VLM and standard flow-matching objective, while replacing only the repeatedly evaluated Transformer action expert with a compact Mamba expert conditioned through compressed VLM context and gated FiLM modulation. We further show that this action-expert replacement transfers to SmolVLA.
\section{Methodology}

Throughout this section, $N$ denotes the number of visual-language prefix tokens, $H$ the action horizon, $d$ the action-expert hidden width, and $d_a$ the padded action dimension. $\mathbf{H}^{(l)}\in\mathbb{R}^{H\times d}$ denotes the action-token sequence after block $l$. Superscript $(l)$ indexes action-expert blocks, whereas $\ell$ denotes the language instruction. Batch dimensions are omitted for clarity.

\subsection{Preliminaries}
\label{subsec:preliminaries}
% \textbf{Flow-matching VLAs and denoising latency bottleneck.}
% The $\pi$ family models ($\pi_0$ and $\pi_{0.5}$) , formulates continuous action generation through flow matching, while lightweight architectures e.g. SmolVLA also adopt the same iterative action generation mechanism. Given a clean action chunk
% $\mathbf{a}_0 \in \mathbb{R}^{H \times d_a}$, Gaussian noise
% $\boldsymbol{\epsilon} \sim \mathcal{N}(0,I)$, and flow time
% $t \sim \mathcal{U}(0,1)$, training constructs an interpolated noisy action chunk as: $    \mathbf{x}_t = t\boldsymbol{\epsilon} + (1-t)\mathbf{a}_0$. The action expert receives $\mathbf{x}_t$, the flow time $t$, and the visual-language observation $\mathbf{o}$, and predicts the corresponding flow velocity
% $\hat{\mathbf{v}}_{\theta}(\mathbf{x}_t,t,\mathbf{o})$.
\textbf{Flow-matching VLAs and denoising latency bottleneck.}
The $\pi$ family of models, including $\pi_0$ and $\pi_{0.5}$, formulates
continuous action generation through flow matching. 
% Lightweight architectures
% such as SmolVLA also adopt an iterative flow-matching action-generation
% procedure.

Let $\mathbf{A}_0\in\mathbb{R}^{H\times d_a}$ be a clean padded action chunk.
Training samples
\begin{equation}
\begin{aligned}
u&\sim\operatorname{Beta}(1.5,1.0),&
t&=0.001+0.999u,\\
\boldsymbol{\epsilon}&\sim\mathcal{N}(\mathbf{0},\mathbf{I}),&
\mathbf{X}_t&=t\boldsymbol{\epsilon}+(1-t)\mathbf{A}_0,
\end{aligned}
\label{eq:flow-matching}
\end{equation}
and the action expert predicts
$\hat{\mathbf{V}}_\theta(\mathbf{X}_t,t,\mathbf{o})$.

% At inference, $\mathbf{X}_1\sim\mathcal{N}(\mathbf{0},\mathbf{I})$, where $\mathbf{I}$ denotes the identity covariance matrix, and $N_{\mathrm{denoise}}$ Euler steps with
% $\Delta t=1/N_{\mathrm{denoise}}$ apply
% \begin{equation}
% \mathbf{X}_{t-\Delta t}
% =
% \mathbf{X}_t-\Delta t\,
% \hat{\mathbf{V}}_\theta(\mathbf{X}_t,t,\mathbf{o}).
% \label{eq:euler}
% \end{equation}
At inference, we initialize
$\mathbf{X}_1\sim\mathcal{N}(\mathbf{0},\mathbf{I})$, where $\mathbf{I}$
denotes the identity covariance matrix and
$\mathbf{X}_t\in\mathbb{R}^{H\times d_a}$ is the noisy action chunk at flow
time $t$. We then integrate the learned flow backward from $t=1$ to $t=0$
using $N_{\mathrm{denoise}}$ Euler denoising steps. With uniform step size
$\Delta t = 1/N_{\mathrm{denoise}}$, each step updates
\begin{equation}
\mathbf{X}_{t-\Delta t}
=
\mathbf{X}_t
-
\Delta t\,
\hat{\mathbf{V}}_\theta(\mathbf{X}_t,t,\mathbf{o}).
\label{eq:euler}
\end{equation}

We use $N_{\mathrm{denoise}}=10$, a common choice in flow-matching VLA
inference~\cite{black2024pi_0,black2025pi_0,shukor2025smolvla}. Although the
visual-language prefix is computed once per policy query, the action expert is
evaluated at every denoising step.

\begin{figure}[!t]
    \centering
    \includegraphics[width=.8\columnwidth,
    trim={3bp 6bp 10bp 25bp},
    clip]{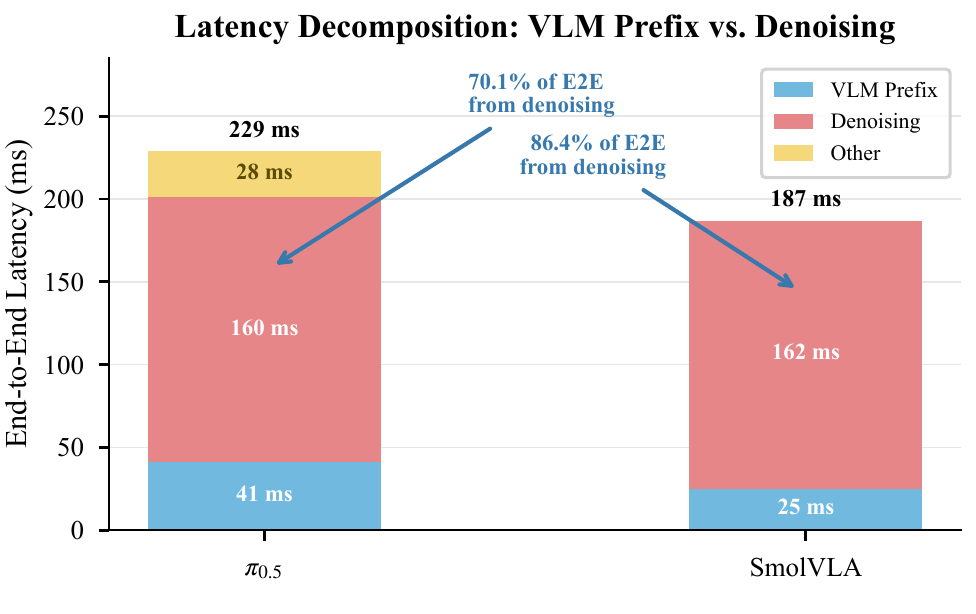}
    \caption{Latency decomposition of $\pi_{0.5}$ and SmolVLA on LIBERO benchmark highlighting action generation as the dominant source of latency.}
    \label{fig:latency-decomposition}
    \vspace{-6mm}
\end{figure}

The action generation cost therefore scales approximately linearly with the number of denoising steps, making the repeatedly evaluated action expert a major inference bottleneck. Figure~\ref{fig:latency-decomposition} confirms this: 10-step denoising accounts for $70.1\%$ of end-to-end latency in $\pi_{0.5}$ and $86.4\%$ in SmolVLA. Thus, iterative action generation remains the dominant cost even with a much smaller visual-language backbone, motivating the quest for a more efficient expert whose per-step savings accumulate across denoising.

\begin{figure*}[!t]
    \centering
    \includegraphics[width=0.67\textwidth]{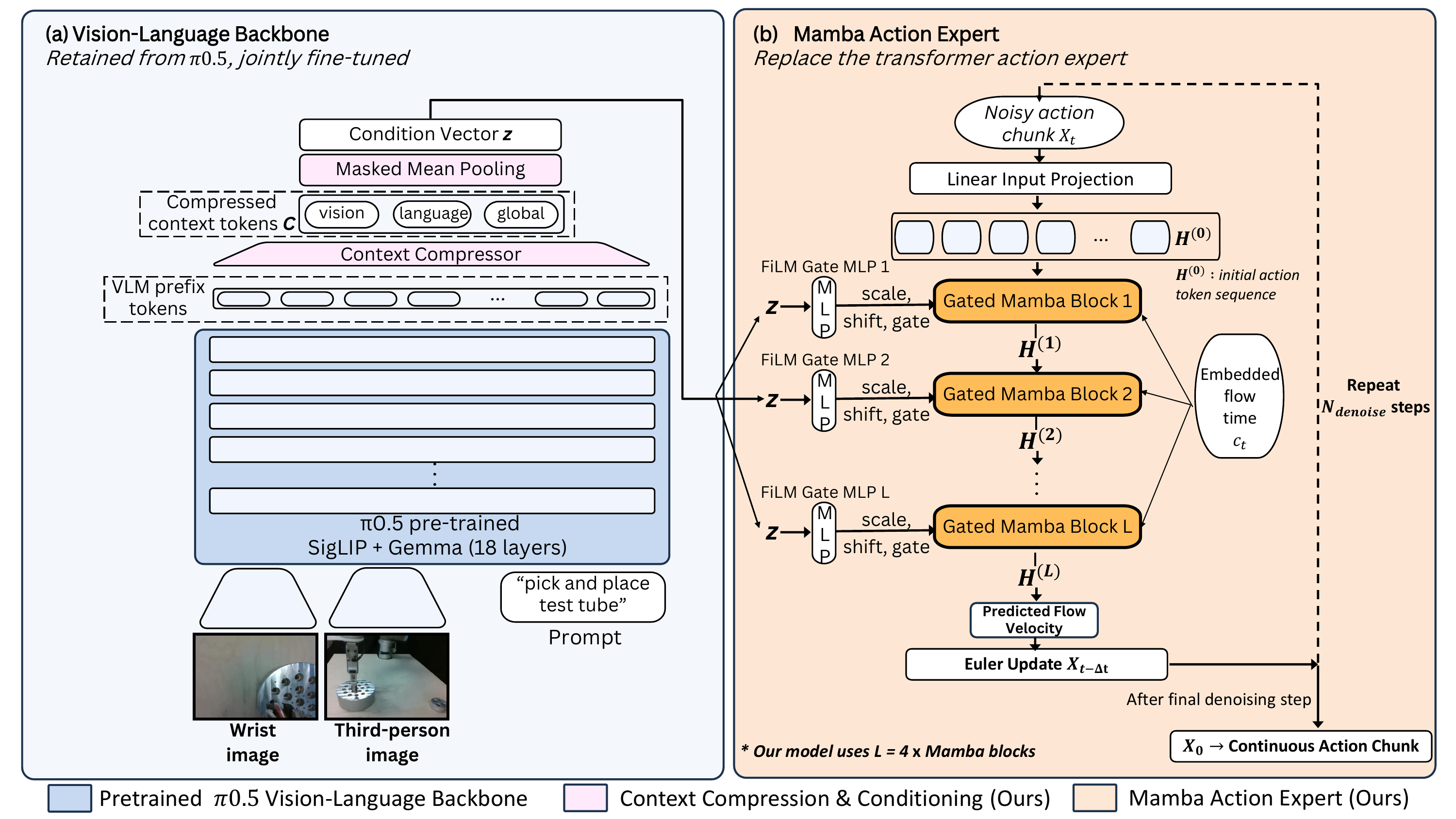}
    \caption{
    \textbf{MambaAct architecture.}
    MambaAct retains and jointly fine-tunes the pretrained $\pi_{0.5}$
    visual-language backbone while replacing its Transformer action expert
    with a compact Mamba-3 expert. The VLM prefix is compressed into visual,
    language, and learned-query global context tokens $\mathbf{C}$ and pooled
    into $\mathbf{z}$, which block-specific FiLM MLPs map to scale, shift, and
    residual-gate parameters. The noisy action chunk $\mathbf{X}_t$ is
    projected into the initial action-token sequence $\mathbf{H}^{(0)}$, while
    the embedded flow time $\mathbf{c}_t$ is injected through block-specific
    projections. The $L=4$ gated Mamba-3 blocks transform the full action-token
    sequence into $\mathbf{H}^{(L)}$, and a final RMSNorm-linear head predicts
    the flow velocity $\hat{\mathbf{V}}_\theta$. Repeated Euler updates use this
    velocity to obtain the denoised action chunk $\mathbf{X}_0$.
    }
    \label{fig:mambaact-architecture}
    \vspace{-6mm}
\end{figure*}

\textbf{Mamba sequence mixers.} We use the SISO Mamba3 formulation~\cite{lahotimamba} along the action-token dimension; its complete selective scan, native gating, skip connection, and internal
mixer projection are denoted by $\operatorname{Mamba3}(\cdot)$.

\subsection{MambaAct}
Figure~\ref{fig:mambaact-architecture} gives an overview of MambaAct. MambaAct retains and jointly fine-tunes the pretrained $\pi_{0.5}$ visual-language backbone, while replacing its $311.5$M-parameter Transformer action expert with a $29.7$M-parameter Mamba expert. Given $\mathbf{o}=\{\mathbf{I}^{w},\mathbf{I}^{e},\ell\}$, where $\mathbf{I}^{w}$ and $\mathbf{I}^{e}$ denote the wrist and third-person images, respectively, and $\ell$ is the language instruction, the backbone produces
\begin{equation}
\mathbf{P}=f_{\mathrm{VLM}}(\mathbf{o})
\in\mathbb{R}^{N\times d_{\mathrm{vlm}}}.
\end{equation}

\paragraph{Context compression}
We split the VLM prefix into visual and language tokens, $\mathbf{P}=[\mathbf{P}_v;\mathbf{P}_\ell]$, where $\mathbf{P}_v$ and $\mathbf{P}_\ell$ denote the visual and language portions of the prefix, respectively. We then project all prefix tokens from the VLM hidden width to the action-expert width using
\begin{equation}
\widetilde{\mathbf{P}}
=
\mathbf{P}\mathbf{W}_c+\mathbf{b}_c
\in\mathbb{R}^{N\times d}.
\end{equation}
where $\mathbf{W}_c$ and $\mathbf{b}_c$ are the learned projection matrix and bias, respectively.
Visual tokens are compressed into $K_v$ contiguous segments:
\begin{equation}
\mathbf{C}_{v,i}
=
\frac{\sum_{n\in\mathcal{B}^v_i}
m^v_n\widetilde{\mathbf{P}}_{v,n}}
{\max(1,\sum_{n\in\mathcal{B}^v_i}m^v_n)},
\qquad i=1,\ldots,K_v,
\end{equation}
where $\mathcal{B}^v_i$ denotes the set of visual-token indices assigned to the $i$-th contiguous segment, $\widetilde{\mathbf{P}}_{v,n}$ is the projected representation of the $n$-th visual token, and $m^v_n\in\{0,1\}$ is its validity mask. Language tokens are pooled analogously into $\mathbf{C}_\ell$. In addition,
$K_g$ learned queries
$\mathbf{Q}_g\in\mathbb{R}^{K_g\times d}$ extract global summaries,
$\mathbf{C}_g=\operatorname{MHA}
(\mathbf{Q}_g,\widetilde{\mathbf{P}},\widetilde{\mathbf{P}})$.
The complete context and condition vectors are
\begin{equation}
\begin{gathered}
\mathbf{C}
=
[\mathbf{C}_v+\mathbf{e}_v;
 \mathbf{C}_\ell+\mathbf{e}_\ell;
 \mathbf{C}_g+\mathbf{e}_g],
\\
\mathbf{C}
\leftarrow
\mathbf{C}
+
\operatorname{MLP}_c
\left(
\operatorname{RMSNorm}(\mathbf{C})
\right),
\\
\mathbf{z}
=
\frac{\sum_{k=1}^{K}m_k\mathbf{C}_k}
     {\max\left(1,\sum_{k=1}^{K}m_k\right)}
\in\mathbb{R}^{d},
\qquad
K=K_v+K_\ell+K_g.
\end{gathered}
\label{eq:context-compression}
\end{equation}
Here, $m_k$ is the context-validity mask, while $\mathbf{e}_v$, $\mathbf{e}_\ell$, and $\mathbf{e}_g$ are modality embeddings for visual, language, and global context tokens, respectively.

\paragraph{Action and flow-time inputs}
Executable actions are padded to $d_a=32$. The noisy chunk is projected into
exactly $H$ action tokens, with no additional robot-state token:
\begin{equation}
\mathbf{H}^{(0)}
=
\mathbf{X}_t\mathbf{W}_{\mathrm{in}}+\mathbf{b}_{\mathrm{in}}
\in\mathbb{R}^{H\times d}.
\label{eq:action-input}
\end{equation}
Here, $\mathbf{H}^{(0)}$ is the initial $H$-token action representation,
$\mathbf{W}_{\mathrm{in}}$ is the learned input projection, and
$\mathbf{b}_{\mathrm{in}}$ its bias.

Flow time is encoded separately as
$\mathbf{c}_t=\operatorname{MLP}_t(\phi(t))$.

% \paragraph{Gated Mamba action expert}
% The expert contains $L=4$ blocks. Each
% $\mathbf{H}^{(l)}\in\mathbb{R}^{H\times d}$ represents the complete
% action-token sequence after block $l$. For $l=0,\ldots,L-1$,
% \begin{equation}
% \begin{aligned}
% \mathbf{q}^{(l)}_t
% &=\mathbf{W}^{(l)}_t\operatorname{SiLU}(\mathbf{c}_t)
%   +\mathbf{b}^{(l)}_t,\\
% \widetilde{\mathbf{H}}^{(l)}
% &=\mathbf{H}^{(l)}
%   +\mathbf{1}_H(\mathbf{q}^{(l)}_t)^\top,\\
% \mathbf{U}^{(l)}
% &=\operatorname{Mamba3}^{(l)}
%   \!\left(\operatorname{RMSNorm}
%   (\widetilde{\mathbf{H}}^{(l)})\right),\\
% (\boldsymbol{\gamma}^{(l)},\boldsymbol{\beta}^{(l)},
%  \boldsymbol{\rho}^{(l)})
% &=\operatorname{MLP}^{(l)}_{\mathrm{gate}}
%   \!\left(\operatorname{RMSNorm}(\mathbf{z})\right),\\
% \overline{\mathbf{U}}^{(l)}
% &=(1+\boldsymbol{\gamma}^{(l)})\odot\mathbf{U}^{(l)}
%   +\boldsymbol{\beta}^{(l)},\\
% \mathbf{R}^{(l)}
% &=\widetilde{\mathbf{H}}^{(l)}
%   +\sigma(\boldsymbol{\rho}^{(l)})
%    \odot\overline{\mathbf{U}}^{(l)},\\
% \mathbf{H}^{(l+1)}
% &=\mathbf{R}^{(l)}
%   +\operatorname{FFN}^{(l)}
%   \!\left(\operatorname{RMSNorm}(\mathbf{R}^{(l)})\right).
% \end{aligned}
% \label{eq:gated-mamba-block}
% \end{equation}

\paragraph{Gated Mamba action expert}
The expert contains $L=4$ blocks. Each
$\mathbf{H}^{(l)}\in\mathbb{R}^{H\times d}$ denotes the action-token sequence
after block $l$. For $l=0,\ldots,L-1$,
\begin{equation}
\setlength{\jot}{1pt}
\begin{aligned}
\mathbf{q}^{(l)}_t
&=\mathbf{W}^{(l)}_t\operatorname{SiLU}(\mathbf{c}_t)
+\mathbf{b}^{(l)}_t,\\
\widetilde{\mathbf{H}}^{(l)}
&=\mathbf{H}^{(l)}
+\mathbf{1}_H(\mathbf{q}^{(l)}_t)^\top,\\
(\boldsymbol{\gamma}^{(l)},\boldsymbol{\beta}^{(l)},
\boldsymbol{\rho}^{(l)})
&=\operatorname{MLP}^{(l)}_{\mathrm{gate}}
\!\left(\operatorname{RMSNorm}(\mathbf{z})\right),\\
\mathbf{U}^{(l)}
&=\operatorname{Mamba3}^{(l)}
\!\left(\operatorname{RMSNorm}
(\widetilde{\mathbf{H}}^{(l)})\right),\\
\overline{\mathbf{U}}^{(l)}
&=(1+\boldsymbol{\gamma}^{(l)})\odot\mathbf{U}^{(l)}
+\boldsymbol{\beta}^{(l)},\\
\mathbf{R}^{(l)}
&=\widetilde{\mathbf{H}}^{(l)}
+\sigma(\boldsymbol{\rho}^{(l)})
\odot\overline{\mathbf{U}}^{(l)},\\
\mathbf{H}^{(l+1)}
&=\mathbf{R}^{(l)}
+\operatorname{FFN}^{(l)}
\!\left(\operatorname{RMSNorm}(\mathbf{R}^{(l)})\right).
\end{aligned}
\label{eq:gated-mamba-block}
\end{equation}

Here, $\mathbf{1}_H\in\mathbb{R}^{H}$ is an all-ones vector that broadcasts
$\mathbf{q}^{(l)}_t$ across the action horizon.
$\mathbf{U}^{(l)}$ denotes the Mamba output,
$\overline{\mathbf{U}}^{(l)}$ its FiLM-modulated representation, and
$\mathbf{R}^{(l)}$ the gated residual state.
$\boldsymbol{\gamma}^{(l)}$, $\boldsymbol{\beta}^{(l)}$, and
$\boldsymbol{\rho}^{(l)}$ are the block-specific FiLM scale, shift, and
residual-gate logits. We zero-initialize the time projections and FiLM
scale/shift outputs, and initialize the gate logits to $\log(0.2/0.8)$.
This gives identity FiLM modulation and an initial residual multiplier of
$0.2$, attenuating the newly initialized Mamba branch while keeping it active.
The value $0.2$ is an initialization hyperparameter rather than a tuned
optimum, and the gates are subsequently learned during fine-tuning.
FiLM parameters are reused across denoising steps, while time conditioning
and action tokens are recomputed at each step.

  \paragraph{Velocity prediction and training}
After the final block, the velocity head and target are
\begin{equation}
\begin{aligned}
\hat{\mathbf{V}}_\theta(\mathbf{X}_t,t,\mathbf{o})
&=\operatorname{RMSNorm}(\mathbf{H}^{(L)})
  \mathbf{W}_{\mathrm{out}}+\mathbf{b}_{\mathrm{out}},\\
\mathbf{V}^{\star}
&=\boldsymbol{\epsilon}-\mathbf{A}_0,
\end{aligned}
\label{eq:velocity-head}
\end{equation}
where $\mathbf{A}_0\in\mathbb{R}^{H\times d_a}$ denotes the clean ground-truth action chunk before noise is added, and $\boldsymbol{\epsilon}$ is the sampled Gaussian noise used to construct $\mathbf{X}_t$. We minimize the
mean flow-matching error
\begin{equation}
\mathcal{L}_{\mathrm{FM}}
=
\mathbb{E}
\left[
\frac{1}{Hd_a}
\left\|
\hat{\mathbf{V}}_\theta-\mathbf{V}^{\star}
\right\|_F^2
\right].
\label{eq:flow-loss}
\end{equation}

% \paragraph{Inference}
% Starting from $\mathbf{X}_1\sim\mathcal{N}(\mathbf{0},\mathbf{I})$, the expert
% predicts a velocity and Eq.~\eqref{eq:euler} updates the noisy chunk at each
% denoising step. After $N_{\mathrm{denoise}}$ steps, the first $d_{\mathrm{env}}$ dimensions
% of $\mathbf{X}_0$ form the executable continuous action chunk, where
% $d_{\mathrm{env}}$ denotes the robot's executable action dimension.

\subsection{Experimental Setup}
We evaluate whether replacing the transformer-based action expert with a lightweight Mamba action expert can preserve or improve policy performance while reducing inference cost. We use LIBERO to measure simulated task performance and latency, LIBERO-Plus to assess zero-shot robustness, and two real-world UR5 tasks - tube placing and nut pouring - to evaluate physical manipulation performance and deployment efficiency.

\paragraph{Base policy and model configuration} We use OpenPI $\pi_{0.5}$~\cite{black2025pi_0} as both the base model and the controlled baseline. MambaAct retains the pretrained PaliGemma VLM backbone~\cite{beyer2024paligemma} including SigLIP vision encoder and Gemma visual-language prefix encoder, while replacing its transformer-based action expert with a lightweight Mamba action expert.

All $\pi_{0.5}$-based variants are initialized from the same base checkpoint and fine-tuned using a single-stage full-model recipe, in which the vision encoder, executed VLM layers, context-compression module, and action expert are optimized jointly. The MambaAct model executes all 18 VLM layers and uses a four-block Mamba action expert with width $512$, state size $64$, expansion factor $1$, and head dimension $64$. The Context Compressor produces $K=64$ context tokens, comprising 44 visual tokens, 16 language tokens, and 4 learned-query global summary tokens.

\paragraph{Training protocol} 
% Unless otherwise stated, models reported in the main experiments are trained for $30{,}000$ optimizer steps with batch size $16$, gradient accumulation $1$, and training seed $42$. Multi-seed validation additionally uses seeds $17$ and $27$. We fine-tune the full VLM backbone together with the full action expert, without freezing either component. We use AdamW with gradient clipping at $1.0$, weight decay $10^{-10}$, and a cosine learning-rate schedule from $5\times10^{-5}$ to $5\times10^{-6}$ without warmup. Training uses bfloat16 precision, gradient checkpointing, eight CPU workers, and a single NVIDIA H100 GPU.
% All policies are trained for $30{,}000$ optimizer steps with batch size $16$, gradient accumulation $1$, and training seed $42$. 

All $\pi_{0.5}$-based policies are trained for $30{,}000$ optimizer steps
with batch size $16$ and gradient accumulation $1$. For the main LIBERO comparison, we independently
fine-tune MambaAct and the matched $\pi_{0.5}$ baseline with training seeds
$\{17,27,42\}$. All other $\pi_{0.5}$-based experiments use seed $42$ unless
otherwise stated. We fine-tune the full VLM backbone together with the full action expert, without freezing either component. We use AdamW with gradient clipping at $1.0$, weight decay $10^{-10}$, and a cosine learning-rate schedule from $5\times10^{-5}$ to $5\times10^{-6}$ without warmup. Training uses bfloat16 precision, gradient checkpointing, eight CPU workers, and a single NVIDIA H100 GPU. 
% For the main LIBERO comparison, we independently fine-tune MambaAct and the matched $\pi_{0.5}$ baseline with fine-tuning seeds $\{17,27,42\}$. All other $\pi_{0.5}$-based experiments use a seed of $42$ unless otherwise stated.

\paragraph{Simulation benchmarks} We evaluate each policy on the standard LIBERO benchmark~\cite{liu2023libero}, including the \textsc{Spatial}, \textsc{Object}, \textsc{Goal}, and \textsc{LIBERO-10} (later denoted as L-10) suites. Each suite contains ten tasks. We run 50 trials per task using evaluation seed $7$ and report the success rate for each suite together with the unweighted average across all four suites. LIBERO inference uses image resolution $224\times224$, action horizon $H=10$, ten flow-matching denoising steps, and replanning every five environment steps.
For the main LIBERO comparison, we report mean and standard deviation across
these three fine-tuning seeds.

However, standard LIBERO performance may mask overfitting to benchmark-specific visual and environmental conditions, as policies can degrade under small perturbations. We therefore evaluate MambaAct and the matched $\pi_{0.5}$ baseline on LIBERO-Plus~\cite{fei2025libero}, which contains $10{,}030$ test-only task instances across seven perturbation categories: CAMERA, ROBOT, LANGUAGE, LIGHT, BACKGROUND, NOISE, and LAYOUT. Following the official protocol, we run one rollout per task instance using a fixed evaluation seed and report category-level and overall success rates.

% For robustness evaluation, we additionally evaluate MambaAct and the matched $\pi_{0.5}$ baseline on LIBERO-Plus~\cite{fei2025libero}. LIBERO-Plus organizes its evaluation variants into seven perturbation categories: \textsc{Camera}, \textsc{Robot}, \textsc{Language}, \textsc{Light}, \textsc{Background}, \textsc{Noise}, and \textsc{Layout}. We run one fixed-seed trial per listed variant and report both category-level and overall success rates.

\begin{figure*}[t]
    \centering
    \includegraphics[
        width= \textwidth
    ]{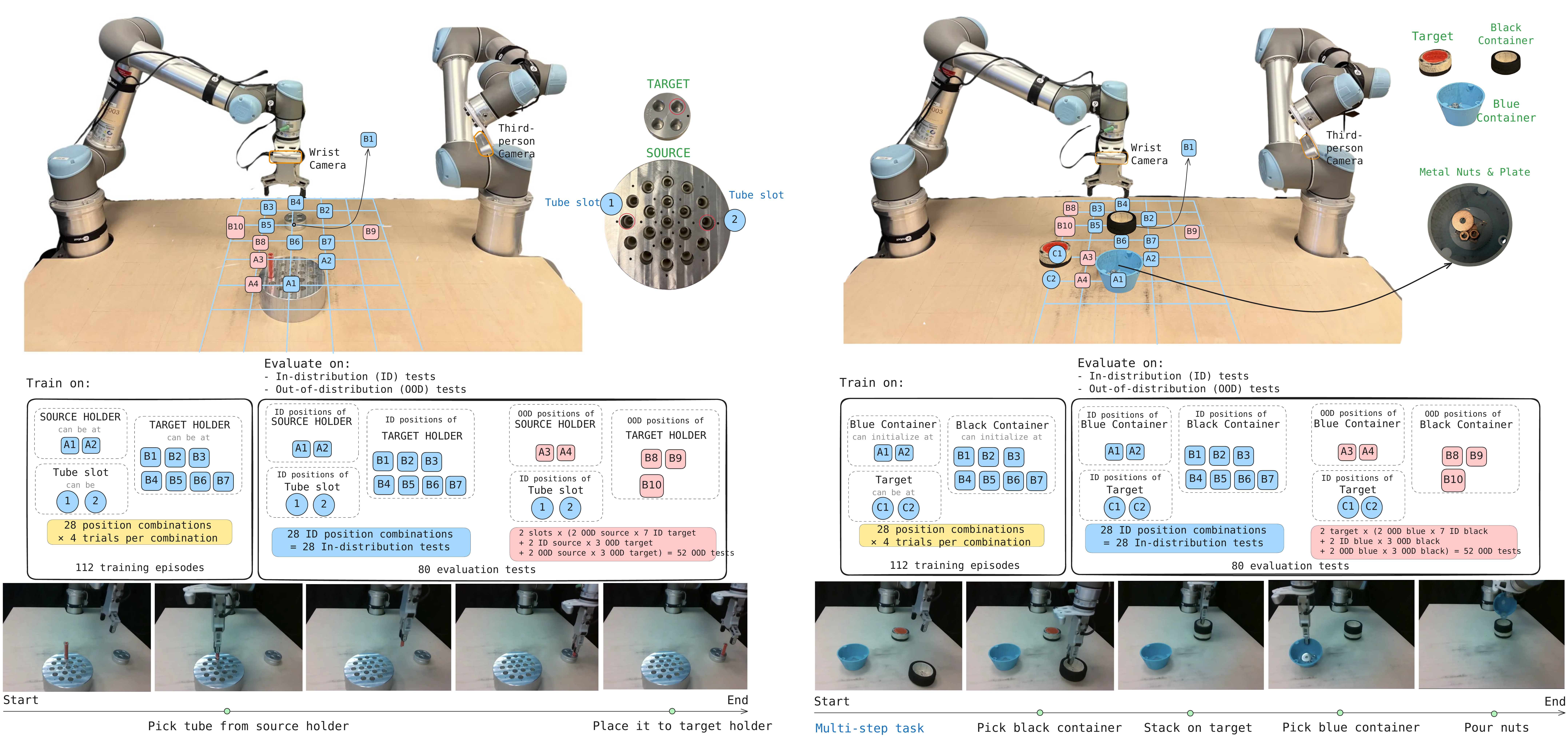}
    \caption{Real-world benchmark setups: (left) tube placing and
    (right) nut pouring.}
    \label{fig:real_tasks}
    \vspace{-4mm}
\end{figure*}

% \paragraph{Real-world benchmarks} Our setup uses two UR5 robots (Fig.~\ref{fig:real_tasks}): the left robot manipulates, while the right serves as a fixed third-person camera mount; a wrist camera provides the second view. Object positions lie on a $10{\times}10$\,cm grid, with blue locations used for training and in-distribution (ID) evaluation and red locations reserved for out-of-distribution (OOD) testing. For each task, we collect 112 demonstrations from 28 training configurations, with four episodes per configuration, and evaluate on 28 ID and 52 OOD trials.
\paragraph{Real-world benchmarks} Our setup uses two UR5 robots (Fig.~\ref{fig:real_tasks}): the left robot performs manipulation and carries a wrist-mounted RealSense camera, while the right robot remains fixed and carries a second RealSense camera providing the third-person view. We use third-person and wrist-mounted RGB observations at $30$~fps and
$640\times480$, resized to $224\times224$ with aspect-ratio-preserving padding. Object positions lie on a $10{\times}10$\,cm grid, with blue locations used for training and in-distribution (ID) evaluation and red locations reserved for out-of-distribution (OOD) testing. For each task, we collect 112 demonstrations from 28 training configurations, with four episodes per configuration, and evaluate on 28 ID and 52 OOD trials.

% \textbf{Tube placing.} 
% The robot transfers a tube from a source to a target holder. Training uses source positions $A1$-$A2$, target positions $B1$-$B7$, and two tube slots; OOD evaluation introduces unseen source positions $A3$-$A4$ and target positions $B8$-$B10$.

% \textbf{Nut pouring.} This longer-horizon task requires the robot to place the black container on the target, grasp the blue container, and pour its contents into the black container.Training uses blue-container positions $A1$-$A2$, black-container positions $B1$-$B7$, and targets $C1$-$C2$; OOD evaluation uses unseen blue-container positions $A3$-$A4$ and black-container positions $B8$-$B10$.
We consider two real-world tasks: \textbf{Tube Placing} and  \textbf{Nut Pouring}. In the former, the robot transfers a tube from a source holder to a target holder; training uses source positions $A1$-$A2$, target positions $B1$-$B7$, and two tube slots, while OOD evaluation introduces unseen source positions $A3$-$A4$ and target positions $B8$-$B10$. In the latter, the robot first places the black container on the target, then grasps the blue container and pours its contents into the black container. Training uses blue-container positions $A1$-$A2$, black-container positions $B1$-$B7$, and targets $C1$-$C2$, while OOD evaluation uses unseen blue-container positions $A3$-$A4$ and black-container positions $B8$-$B10$.

% \begin{figure}[!t]
%     \centering
%     \includegraphics[width=\columnwidth]{figures/full_tube.pdf}
%     \caption{Real-world tube-placing benchmark setup.}
%     \label{fig:tube-benchmark}
% \end{figure}

% \begin{figure}[!t]
%     \centering
%     \includegraphics[width=\columnwidth]{figures/full_pour.pdf}
%     \caption{Real-world nut-pouring benchmark setup.}
%     \label{fig:pour-benchmark}
% \end{figure}

% \begin{figure*}[t]
%     \centering
%     \includegraphics[
%         width=\textwidth,
%         trim={36bp 305bp 41bp 0bp},
%         clip
%     ]{figures/real_tasks.pdf}
%     \caption{Real-world benchmark setups: (left) tube placing and
%     (right) nut pouring.}
%     \label{fig:real_tasks}
% \end{figure*}

For real-world evaluation, we use a normalized success rate with partial credit for meaningful intermediate completion. For tube placing, a trial receives $1$ for successful placement, $0.5$ when the tube is grasped but not inserted into the target holder, and $0$ when grasping fails. For nut pouring, a trial receives $1$ for successful pouring, $0.5$ when the black container is successfully placed but pouring fails, and $0$ when the preceding grasp-and-place stage fails.

\paragraph{Real-world inference}
For physical deployment, the policies predict delta-joint action chunks with horizon $H=50$ from the visual-language observation. At each policy query, the robot executes the first 30 predicted actions at a control frequency of $30\,\mathrm{Hz}$ before replanning. Real-world inference runs on an NVIDIA RTX 5090 GPU.

\paragraph{Latency measurement}
We report \emph{Prefix}, \emph{Denoise}, and \emph{E2E} (end-to-end) latency.
Prefix covers visual-language encoding and context construction, Denoise covers
all ten flow-matching steps, and \emph{E2E} includes the full policy query with I/O,
CPU-GPU transfer, serialization, communication, and scheduling overhead.
Simulation and real-world latency are measured on a single H100 and RTX~5090,
respectively.

% \section{Experiments and Results}
\section{Results}

\subsection{Main Results}
\subsubsection{LIBERO}

% Table~\ref{tab:libero-results} highlights the performance and efficiency gains of MambaAct on LIBERO. MambaAct improves the average success rate from $96.5\%$ to $98.4\%$, while reducing the action-expert core from $311.5$M to $29.7$M parameters ($\downarrow 90.5\%$). Replacing the original Gemma-style action pathway also removes its associated action-side parameter tensors; after adding the lightweight Mamba blocks, Context Compressor, gating modules, and projection heads, the total model size decreases from $3.62$B to $2.96$B parameters. MambaAct also delivers substantial inference speedups. While the Context Compressor adds modest prefix overhead ($40.76$ to $46.61$~ms), it reduces repeated denoising latency by $64.8\%$, resulting in a $52.9\%$ reduction in E2E latency. 
Table~\ref{tab:libero-results} highlights the performance and efficiency gains of MambaAct on LIBERO. Across three independent fine-tuning seeds, MambaAct improves average success from $96.72\pm0.14\%$ to $97.80\pm0.43\%$ ($+1.08$ percentage points), while reducing the action-expert core from $311.5$M to $29.7$M parameters ($\downarrow90.5\%$). Replacing the original Gemma-style action pathway also removes its associated action-side parameter tensors, reducing the total model size from $3.62$B to $2.96$B parameters after introducing the Mamba expert and conditioning modules. On the seed-$42$ checkpoints used for latency measurement, the Context Compressor adds modest prefix overhead ($40.76$ to $46.61$~ms), while denoising latency decreases by $64.8\%$, yielding a $52.9\%$ reduction in E2E latency.

% \begin{table}[!htbp]
% \begin{table}[!t]
%     \centering
%     \scriptsize
%     \caption{LIBERO performance and efficiency results measured on a single H100. Success rates are reported in percent. Latency is decomposed into VLM prefix encoding and flow-matching action denoising.}
%     \label{tab:libero-results}
%     \setlength{\tabcolsep}{1.5pt}
%     \renewcommand{\arraystretch}{1.08}
%     \resizebox{\columnwidth}{!}{%
%     \begin{tabular}{@{}lccccccccc@{}}
%         \toprule
%         & & & & & & & \multicolumn{3}{c}{Latency (ms)} \\
%         \cmidrule(lr){8-10}
%         Model & Spatial & Object & Goal & L-10 & Avg. & \shortstack{Model Size\\(B)} & Prefix & Denoise & E2E \\
%         \midrule
%         $\pi_{0.5}$ & $97.6$ & $98.8$ & $96.6$ & $93.2$ & $96.5$ & $3.62$ & $40.76$ & $160.26$ & $228.6$ \\
%         \cellcolor{Green!15}MambaAct & \cellcolor{Green!15}$\textbf{99.2}$ & \cellcolor{Green!15}$\textbf{99.8}$ & \cellcolor{Green!15}$\textbf{98.8}$ & \cellcolor{Green!15}$\textbf{95.8}$ & \cellcolor{Green!15}$\textbf{98.4}$ & \cellcolor{Green!15}$2.96$ & \cellcolor{Green!15}$46.61$ & \cellcolor{Green!15}$56.40$ {\color{green!50!black}$\downarrow64.8\%$} & \cellcolor{Green!15}$107.6$ {\color{green!50!black}$\downarrow52.9\%$} \\
%         \bottomrule
%     \end{tabular}%
%     }
%     \vspace{-5mm}
% \end{table}
\begin{table}[!t]
    \centering
    \scriptsize
    \caption{LIBERO performance and efficiency results. Success rates are
    mean$\pm$standard deviation over three independent fine-tuning seeds
    $\{17,27,42\}$, with $50$ trials per task using evaluation seed $7$.
    Latency is measured on the seed-$42$ checkpoints using a single H100.}
    \label{tab:libero-results}

    \setlength{\tabcolsep}{2.0pt}
    \renewcommand{\arraystretch}{1.08}

    % Success-rate table
    \resizebox{0.95\columnwidth}{!}{%
    \begin{tabular}{@{}lccccc@{}}
        \toprule
        \multicolumn{6}{c}{\textbf{LIBERO success rate (\%)}} \\
        \midrule
        Model & Spatial & Object & Goal & L-10 & Avg. \\
        \midrule

        $\pi_{0.5}$
        & $98.00\pm0.43$
        & $98.33\pm0.66$
        & $97.73\pm0.84$
        & $92.80\pm0.43$
        & $96.72\pm0.14$ \\

        \rowcolor{Green!15}
        MambaAct
        & $\mathbf{98.73\pm0.34}$
        & $\mathbf{99.60\pm0.43}$
        & $\mathbf{98.60\pm0.28}$
        & $\mathbf{94.27\pm1.52}$
        & $\mathbf{97.80\pm0.43}$ \\

        \bottomrule
    \end{tabular}%
    }

    \vspace{1mm}

    % Efficiency and latency table
    \resizebox{0.7\columnwidth}{!}{%
    \begin{tabular}{@{}lcccc@{}}
        \toprule
        \multicolumn{5}{c}{\textbf{Model efficiency and latency}} \\
        \midrule
        &
        &
        \multicolumn{3}{c}{Latency (ms)} \\
        \cmidrule(lr){3-5}
        Model
        & \shortstack{Model Size\\(B)}
        & Prefix
        & Denoise
        & E2E \\
        \midrule

        $\pi_{0.5}$
        & $3.62$
        & $40.76$
        & $160.26$
        & $228.6$ \\

        \rowcolor{Green!15}
        MambaAct
        & $2.96$
        & $46.61$
        & $56.40$ {\color{green!50!black}$\downarrow64.8\%$}
        & $107.6$ {\color{green!50!black}$\downarrow52.9\%$} \\

        \bottomrule
    \end{tabular}%
    }

    \vspace{-5mm}
\end{table}

\begin{table}[!t]
    \centering
    \caption{LIBERO comparison with efficient VLA methods. Speedup is measured
    relative to the listed baseline. Bold and underlined values denote the
    highest and second-highest values listed, respectively. For $\pi_{0.5}$ and
    MambaAct, success rates are means over three independent fine-tuning seeds
    $\{17,27,42\}$; other results are from the cited works.}
    \label{tab:libero-comparison}
    \scriptsize
    \setlength{\tabcolsep}{1.2pt}
    \renewcommand{\arraystretch}{0.92}

    \resizebox{\columnwidth}{!}{%
    \begin{tabular}{@{}p{2.65cm}p{1.85cm}cccccc@{}}
        \toprule
        Method & Baseline & Spatial & Object & Goal & L-10 & Avg. & Speedup \\
        \midrule

        \rowcolor{gray!15}
        \multicolumn{8}{@{}l@{}}{\textbf{Controlled baseline}} \\
        $\pi_{0.5}$~\cite{black2025pi_0}
        & -- & $98.0$ & $98.3$ & $97.7$ & $92.8$ & $96.7$ & $1.00\times$ \\

        \rowcolor{gray!15}
        \multicolumn{8}{@{}l@{}}{\textbf{Action-expert replacement \& conditioning}} \\
        \rowcolor{blue!12}
        MambaAct (Ours)
        & $\pi_{0.5}$
        & $\underline{98.7}$
        & $\underline{99.6}$
        & $\underline{98.6}$
        & $94.3$
        & $\underline{97.8}$
        & $\underline{2.12\times}$ \\

        \rowcolor{gray!15}
        \multicolumn{8}{@{}l@{}}{\textbf{Denoising-step compression}} \\
        SnapFlow~\cite{luan2026snapflow}
        & $\pi_{0.5}$
        & $\mathbf{99.0}$
        & $\mathbf{100.0}$
        & $\mathbf{99.0}$
        & $\mathbf{97.0}$
        & $\mathbf{98.8}$
        & $\mathbf{3.30\times}$ \\

        \rowcolor{gray!15}
        \multicolumn{8}{@{}l@{}}{\textbf{VLM/action-head depth distillation}} \\
        Shallow-$\pi$ ($\pi_{0.5}$-L9)~\cite{jeon2026shallow}
        & $\pi_{0.5}$
        & $\mathbf{99.0}$
        & $98.0$
        & $97.0$
        & $93.0$
        & $97.0$
        & $1.72\times$ \\

        \rowcolor{gray!15}
        \multicolumn{8}{@{}l@{}}{\textbf{Structural layer pruning}} \\
        $\pi_0$-CLP~\cite{nguyen2026finetuning}
        & $\pi_0$
        & $95.0$ & $99.2$ & $95.0$ & $86.4$ & $93.9$ & $1.39\times$ \\

        GR00T-N1.5-CLP~\cite{nguyen2026finetuning}
        & GR00T-N1.5
        & $89.4$ & $98.8$ & $95.8$ & $88.6$ & $93.0$ & $1.42\times$ \\

        \rowcolor{gray!15}
        \multicolumn{8}{@{}l@{}}{\textbf{Visual-token pruning}} \\
        LightVLA~\cite{jiang2025better}
        & OpenVLA-OFT
        & $98.4$ & $98.4$ & $98.2$ & $94.6$ & $97.4$ & n/a \\

        ADP~\cite{pei2025action}
        & OpenVLA-OFT
        & $97.6$ & $98.4$ & $97.4$ & $84.2$ & $94.4$ & $1.35\times$ \\

        $\pi_0$-SpecPrune-VLA~\cite{wang2025specprune}
        & $\pi_0$
        & $96.6$ & $98.0$ & $95.2$ & $84.2$ & $93.5$ & $1.31\times$ \\

        FastV~\cite{chen2024image}
        & OpenVLA-OFT
        & $94.6$ & $95.8$ & $94.0$ & $88.8$ & $93.3$ & $1.44\times$ \\

        DivPrune~\cite{alvar2025divprune}
        & OpenVLA-OFT
        & $92.4$ & $91.2$ & $89.0$ & $84.8$ & $89.4$ & $1.46\times$ \\

        \rowcolor{gray!15}
        \multicolumn{8}{@{}l@{}}{\textbf{Routing and sparsification}} \\
        CogVLA~\cite{li2026cogvla}
        & OpenVLA
        & $\mathbf{99.0}$
        & $99.0$
        & $97.0$
        & $\underline{95.0}$
        & $97.0$
        & n/a \\

        \rowcolor{gray!15}
        \multicolumn{8}{@{}l@{}}{\textbf{Hybrid multi-axis acceleration}} \\
        EfficientVLA~\cite{yang2026efficientvla}
        & OpenVLA-OFT
        & $96.5$ & $91.1$ & $96.0$ & $72.1$ & $88.9$ & $1.52\times$ \\

        \rowcolor{gray!15}
        \multicolumn{8}{@{}l@{}}{\textbf{Compact or alternative architectures}} \\
        SUREFlow~\cite{islam2026sureflow}
        & Compact Mamba
        & $94.8$ & $91.0$ & $93.8$ & $90.2$ & $92.5$ & n/a \\

        SmolVLA~\cite{shukor2025smolvla}
        & Compact VLA
        & $90.0$ & $96.0$ & $92.0$ & $71.0$ & $87.3$ & n/a \\

        OpenVLA~\cite{kim2024openvla}
        & Autoregressive
        & $84.9$ & $88.4$ & $79.2$ & $53.7$ & $76.6$ & n/a \\

        \bottomrule
    \end{tabular}%
    }
    \vspace{-3mm}
\end{table}
% Table~\ref{tab:libero-comparison} compares MambaAct with recent efficient VLA methods on LIBERO. Since absolute inference speeds are reported under heterogeneous hardware and implementation settings, we use speedup relative to each method's own baseline as the primary efficiency metric. MambaAct achieves the second-highest average success in the table, trailing only SnapFlow, while using $50$ trials per task ($2{,}000$ episodes) compared with SnapFlow's $10$ trials per task ($400$ episodes). Unlike SnapFlow, which uses an additional progressive self-distillation stage, MambaAct is trained directly within the standard downstream fine-tuning procedure. It therefore provides a complementary efficiency axis, as further demonstrated by combining MambaAct with VLM-depth reduction..
Table~\ref{tab:libero-comparison} compares MambaAct with recent efficient VLA methods on LIBERO. Given heterogeneous hardware and implementation settings, we report speedup relative to each method's own baseline. MambaAct achieves $97.80\pm0.43\%$ average success across three independent fine-tuning seeds, the second-highest average listed. Its best run (seed $42$) reaches $99.2\%$, $99.8\%$, $98.8\%$, and $95.8\%$ on Spatial, Object, Goal, and L-10, respectively, placing within the top two on every suite and averaging $98.4\%$, close to SnapFlow's reported $98.8\%$. SnapFlow~\cite{luan2026snapflow}, however, evaluates $10$ trials per task and does not report variability across fine-tuning seeds, whereas we evaluate three independently fine-tuned checkpoints with $50$ trials per task each. Moreover, SnapFlow retains the full $\pi_{0.5}$ architecture while compressing its denoising trajectory to one step; in contrast, MambaAct reduces the action-expert core from $311.5$M to $29.7$M parameters while retaining standard flow-matching inference. SnapFlow also uses progressive self-distillation, whereas MambaAct is trained within standard downstream fine-tuning without a separate distillation stage. Beyond simulation, we additionally validate MambaAct on two real-world UR5 manipulation tasks, and release our code and model weights to support reproducibility.

% It therefore provides a lightweight and complementary efficiency upgrade that can also be combined with other acceleration strategies.
% The complementarity labels have the following meanings: \textbf{Yes} denotes methods that are mostly composable with MambaAct, such as SnapFlow or token pruning; \textbf{No} denotes baselines or alternative architectures that are not directly composable; and \textbf{Ours} denotes our MambaAct result.

% \textbf{Multi-seed validation.}
% To assess training variability, we additionally fine-tune MambaAct and the
% matched $\pi_{0.5}$ baseline with two extra training seeds, $17$ and $27$,
% in addition to the original seed $42$. Across the three seeds
% $\{17,27,42\}$, MambaAct outperforms the baseline in every run, achieving
% $97.80\pm0.43\%$ average LIBERO success compared with
% $96.72\pm0.14\%$ for $\pi_{0.5}$, corresponding to a $1.08$
% percentage-point improvement on average.

\subsubsection{LIBERO-Plus} 

\begin{table}[t]
    \centering
    \scriptsize
    \setlength{\tabcolsep}{3pt}
    \renewcommand{\arraystretch}{1.05}
    \caption{
    Zero-shot robustness on LIBERO-Plus. All models are trained on standard LIBERO and evaluated without LIBERO-Plus fine-tuning. Published results are from prior work~\cite{fei2025libero}; $\pi_{0.5}$ and MambaAct use our matched training and evaluation protocol.
    }
    \label{tab:libero_plus}
    \resizebox{0.48\textwidth}{!}{%
    \begin{tabular}{@{}lcccccccc@{}}
        \toprule
        \textbf{Method} & \textbf{Camera} & \textbf{Robot} & \textbf{Language} & \textbf{Light} & \textbf{Backgr.} & \textbf{Noise} & \textbf{Layout} & \textbf{Total} \\
        \midrule
        OpenVLA & 0.8 & 3.5 & 23.0 & 8.1 & 34.8 & 15.2 & 28.5 & 15.6 \\
        OpenVLA-OFT & 56.4 & 31.9 & 79.5 & 88.7 & 93.3 & 75.8 & 74.2 & 69.6 \\
        OpenVLA-OFT$_w$ & 10.4 & 38.7 & 70.5 & 76.8 & \textbf{93.6} & 49.9 & 69.9 & 55.8 \\
        OpenVLA-OFT$_m$ & 55.6 & 21.7 & \textbf{81.0} & 92.7 & 91.0 & 78.6 & 68.7 & 67.9 \\
        NORA & 2.2 & 37.0 & 65.1 & 45.7 & 58.6 & 12.8 & 62.1 & 39.0 \\
        WorldVLA & 0.1 & 27.9 & 41.6 & 43.7 & 17.1 & 10.9 & 38.0 & 25.0 \\
        UniVLA & 1.8 & 46.2 & 69.6 & 69.0 & 81.0 & 21.2 & 31.9 & 42.9 \\
        $\pi_0$ & 13.8 & 6.0 & 58.8 & 85.0 & 81.4 & \textbf{79.0} & 68.9 & 53.6 \\
        $\pi_0$-Fast & \textbf{65.1} & 21.6 & 61.0 & 73.2 & 73.2 & 74.4 & 68.8 & 61.6 \\
        RIPT-VLA & 55.2 & 31.2 & 77.6 & 88.4 & 91.6 & 73.5 & 74.2 & 68.4 \\
        \midrule
        $\pi_{0.5}$ & 61.6 & 70.9 & 76.4 & 94.9 & 92.7 & 77.4 & 86.4 & 78.7 \\
        \rowcolor{blue!12}\textbf{MambaAct (Ours)} & 61.5 & \textbf{77.6} & 79.8 & \textbf{95.2} & 93.2 & 73.5 & \textbf{86.4} & \textbf{79.7} \\
        \bottomrule
    \end{tabular}%
    }
    \vspace{-5mm}
\end{table}

Table~\ref{tab:libero_plus} shows that MambaAct achieves the highest overall score at $79.7\%$, slightly outperforming the matched $\pi_{0.5}$ baseline at $78.7\%$. The largest gain is observed under variations in the robot's initial pose,
where MambaAct improves from $70.9\%$ to $77.6\%$, while MambaAct also achieves the highest listed lighting robustness score at $95.2\%$ and matches the strongest layout result at $86.4\%$. The main degradation occurs under observation noise, where performance drops from $77.4\%$ to $73.5\%$, while camera robustness is essentially unchanged. Overall, the results indicate that the strong standard LIBERO performance of MambaAct does not come at the cost of substantially reduced robustness to perturbations.

\subsubsection{Real-world benchmarks}
Results are shown in Table~\ref{tab:real_world_results}. Both models achieve $100\%$ ID success, while the main difference appears under OOD evaluation. Compared with the baseline  $\pi_{0.5}$, MambaAct improves OOD success from $86.5\%$ to $95.2\%$ on nut pouring and from $76.9\%$ to $82.7\%$ on tube placing, while also reducing model size, GFLOPs, and training time. Most notably, denoising latency decreases by $58.8\%$, yielding roughly $33\%$ lower end-to-end latency on both tasks. Overall, MambaAct improves real-world OOD robustness while substantially accelerating action generation.

\begin{table}[!t]
    \centering
    \caption{
        Real-world efficiency and success rates for nut pouring and tube placing,
        with inference run on an RTX5090.
        E2E latency is the mean end-to-end policy-query latency and includes
        prefix encoding, action denoising, input/output transformations,
        CPU-GPU transfers, serialization, WebSocket communication, and runtime scheduling.
    }
    \label{tab:real_world_results}
    \setlength{\tabcolsep}{4pt}
    \renewcommand{\arraystretch}{1.05}

    % ---- Block 1: model footprint ----
    \resizebox{0.75\columnwidth}{!}{%
    \begin{tabular}{@{}llccc@{}}
        \toprule
        Task & Model & Model Size (B) & Train (h) & GFLOPs \\
        \midrule
        \multirow{2}{*}{Nut pouring}
        & Baseline $\pi_{0.5}$ & $3.62$ & $14.3$ & $5001.2$ \\
        & \cellcolor{Green!15}MambaAct
          & \cellcolor{Green!15}$2.96$ {\color{green!50!black}\scriptsize$\downarrow18.2\%$}
          & \cellcolor{Green!15}$14.0$
          & \cellcolor{Green!15}$4488.8$ {\color{green!50!black}\scriptsize$\downarrow10.2\%$} \\
        \midrule
        \multirow{2}{*}{Tube placing}
        & Baseline $\pi_{0.5}$ & $3.62$ & $14.3$ & $5001.2$ \\
        & \cellcolor{Green!15}MambaAct
          & \cellcolor{Green!15}$2.96$ {\color{green!50!black}\scriptsize$\downarrow18.2\%$}
          & \cellcolor{Green!15}$14.0$
          & \cellcolor{Green!15}$4488.8$ {\color{green!50!black}\scriptsize$\downarrow10.2\%$} \\
        \bottomrule
    \end{tabular}%
    }

    \vspace{2mm}

    % ---- Block 2: latency & success rate ----
    \resizebox{\columnwidth}{!}{%
    \begin{tabular}{@{}llccccccccc@{}}
        \toprule
        & &
        \multicolumn{3}{c}{Latency (ms)} &
        \multicolumn{3}{c}{Success rate (\%)} \\
        \cmidrule(lr){3-5}
        \cmidrule(lr){6-8}
        Task & Model & Prefix & Denoise & E2E & ID & OOD & Final \\
        \midrule
        \multirow{2}{*}{Nut pouring}
        & Baseline $\pi_{0.5}$ & $39.45$ & $74.82$ & $125.49$ & $100.0$ & $86.5$ & $91.3$ \\
        & \cellcolor{Green!15}MambaAct
          & \cellcolor{Green!15}$41.03$
          & \cellcolor{Green!15}$30.84$ {\color{green!50!black}\scriptsize$\downarrow58.8\%$}
          & \cellcolor{Green!15}$82.97$ {\color{green!50!black}\scriptsize$\downarrow33.9\%$}
          & \cellcolor{Green!15}$100.0$
          & \cellcolor{Green!15}$95.2$
          & \cellcolor{Green!15}$96.9$ \\
        \midrule
        \multirow{2}{*}{Tube placing}
        & Baseline $\pi_{0.5}$ & $39.12$ & $75.42$ & $124.84$ & $100.0$ & $76.9$ & $85.0$ \\
        & \cellcolor{Green!15}MambaAct
          & \cellcolor{Green!15}$41.15$
          & \cellcolor{Green!15}$31.11$ {\color{green!50!black}\scriptsize$\downarrow58.8\%$}
          & \cellcolor{Green!15}$83.26$ {\color{green!50!black}\scriptsize$\downarrow33.3\%$}
          & \cellcolor{Green!15}$100.0$
          & \cellcolor{Green!15}$82.7$
          & \cellcolor{Green!15}$88.8$ \\
        \bottomrule
    \end{tabular}%
    }
    \vspace{-5mm}
\end{table}
\subsection{Ablation Studies}

We conduct three ablation studies to assess the source of gain, composability, and transferability of MambaAct. 
Unless otherwise stated, all ablations use a fine-tuning seed of $42$ and an evaluation
seed of $7$.

\textbf{Ablation 1: Architectural source of gain. }We isolate the contributions of the action mixer and conditioning interface using a $2\times2$ comparison of Mamba and Transformer mixers with gated and cross-attention conditioning (Table~\ref{tab:context_mixer_ablation}).
% We then test composability with VLM-depth reduction and transferability to a second flow-matching VLA, SmolVLA \cite{shukor2025smolvla}.

Under gated conditioning, Mamba achieves $98.4\%$ average success, substantially outperforming the size-matched Transformer at $85.7\%$. The same trend holds with cross-attention ($96.7\%$ vs. $75.6\%$), suggesting that the Mamba action mixer is important for preserving performance under aggressive action-expert compression. We hypothesize that selective state-space mixing provides a useful inductive bias for flow-matching action generation: Mamba propagates information along the ordered, strongly correlated action-token sequence through a selective recurrent state rather than unrestricted pairwise token interactions. This may allow a compact expert to model action-token dynamics more effectively, consistent with prior results on Mamba-based diffusion policies~\cite{cao2025mamba}.

Gated conditioning also consistently outperforms cross-attention: with Mamba, it improves average success from $96.7\%$ to $98.4\%$ while reducing latency from $119.66$ to $107.61$ ms; with Transformer, it improves success from $75.6\%$ to $85.7\%$. These results suggest that repeated token-level attention to the compressed context is not necessary for strong action generation in the evaluated setting. However, the gated models use larger action experts ($30$M vs. $17$-$18$M parameters), so the improvement cannot be attributed to conditioning alone.

One possible explanation is the optimization setting of MambaAct. The action expert and conditioning interface are initialized from scratch, while the VLM starts from pretrained $\pi_{0.5}$ weights and is jointly fine-tuned. We hypothesize that gated FiLM provides a constrained interface for aligning these components: feature-wise scale and shift adapt the action representation to the visual-language condition, while the residual gate controls the magnitude of the conditioned update. This interpretation is consistent with prior findings on FiLM~\cite{perez2018film} and adaptive modulation in diffusion transformers~\cite{peebles2023scalable}. Overall, the ablation suggests that Mamba provides a stronger performance-capacity trade-off under aggressive action-expert compression. Gated variants further improve the observed efficiency-performance trade-off, although the accuracy contribution of the conditioning interface remains partially confounded by expert size.  
% consistent with FiLM~\cite{perez2018film} and with conditioning studies in Diffusion Transformer~\cite{peebles2023scalable}, where adaptive scale-shift modulation with gated residual updates was found to be an effective alternative to cross-attention. Although these results do not establish the mechanism in MambaAct, they provide supporting evidence for the effectiveness of lightweight modulation when conditioning a newly initialized generative module.
\begin{table}[!t]
    \centering
    \caption{Action-mixer and conditioning-interface ablation on LIBERO. All models use the same Context Compressor. Size reports action-expert core size.}
    \label{tab:context_mixer_ablation}
    \scriptsize
    \setlength{\tabcolsep}{2.4pt}
    \renewcommand{\arraystretch}{1.08}
    \resizebox{\columnwidth}{!}{%
    \begin{tabular}{@{}llccccc ccc@{}}
        \toprule
        Mixer & Conditioning & Spatial & Object & Goal & L-10 & Avg 
        & Size (B) & Train (h) & Latency (ms) \\
        \midrule
        \cellcolor{blue!12}Mamba 
            & \cellcolor{blue!12}Gate 
            & \cellcolor{blue!12}99.2 
            & \cellcolor{blue!12}99.8 
            & \cellcolor{blue!12}98.8 
            & \cellcolor{blue!12}95.8 
            & \cellcolor{blue!12}\textbf{98.4} 
            & \cellcolor{blue!12}0.030 
            & \cellcolor{blue!12}13.78 
            & \cellcolor{blue!12}107.61 \\
        Mamba 
            & Cross-Attn 
            & 98.6 & 99.4 & 94.6 & 94.2 & 96.7 
            & 0.017 & 13.83 & 119.66 \\
        Transformer 
            & Gate 
            & 82.8 & 90.6 & 86.8 & 82.6 & 85.7 
            & 0.030 & 13.69 & 74.22 \\
        Transformer 
            & Cross-Attn 
            & 75.2 & 82.8 & 75.8 & 68.6 & 75.6 
            & 0.018 & 13.72 & 85.60 \\
        \bottomrule
    \end{tabular}%
    }
    \vspace{-5mm}
\end{table}

% \textbf{Ablation 2: Composability.} Here, we study whether MambaAct can compose with complementary prefix-side VLA acceleration. Efficient VLA systems can be accelerated along multiple axes, including reducing denoising steps, pruning or compressing visual tokens, truncating VLM depth, and replacing the action expert. Because MambaAct primarily reduces computation in the repeatedly evaluated action pathway, its acceleration mechanism is distinct from methods that reduce computation in the visual-language prefix. We test this composability using VLM-depth reduction as a representative prefix-side acceleration. We combine MambaAct with VLM-depth reduction by executing only the first $12$ VLM layers instead of all $18$ layers before passing the resulting context representation to the Mamba action expert.
\textbf{Ablation 2: Composability.} Here, we study whether MambaAct can compose with complementary prefix-side VLA acceleration. Efficient VLA systems can be accelerated along multiple axes, including reducing denoising steps, pruning or compressing visual tokens, truncating VLM depth, and replacing the action expert. Because MambaAct primarily reduces computation in the repeatedly evaluated action pathway, its acceleration mechanism is distinct from methods that reduce computation in the visual-language prefix. We test this composability using VLM-depth reduction as a representative prefix-side acceleration. As a controlled efficiency--accuracy trade-off, we combine MambaAct with VLM-depth reduction by executing only the first $12$ VLM layers instead of all $18$ layers before passing the resulting context representation to the Mamba action expert.

Tables~\ref{tab:libero-mambaact-depth} and~\ref{tab:ablation_real_world} show that MambaAct remains effective when combined with prefix-side acceleration. On LIBERO, the L12 variant reduces training time from $13.78$ to $10.81$ hours and lowers E2E latency from $107.6$ to $102.91$ ms, while maintaining $97.5\%$ average success compared with $98.4\%$ for the full model. This suggests that the Mamba action expert can operate effectively on intermediate VLM representations, rather than requiring the full-depth visual-language prefix.
\begin{table}[!t]
    \centering
    \caption{MambaAct composed with VLM-depth reduction on LIBERO. Both variants use fine-tuning seed $42$ and are evaluated with $50$ trials per task using evaluation seed $7$. Latency is measured on a single H100.}   \label{tab:libero-mambaact-depth}
    \scriptsize
    \setlength{\tabcolsep}{1.2pt}
    \renewcommand{\arraystretch}{0.95}
    \resizebox{0.8\columnwidth}{!}{%
    \begin{tabular}{@{}lcccccccccc@{}}
        \toprule
        & & & & & & & \multicolumn{3}{c}{Latency (ms)} \\
        \cmidrule(lr){8-10}
        Model & Spatial & Object & Goal & L-10 & Avg. & \shortstack{Train\\(h)} & Prefix & Denoise & E2E \\
        \midrule
        MambaAct & $99.2$ & $99.8$ & $98.8$ & $95.8$ & $98.4$ & $13.78$ & $46.61$ & $56.40$ & $107.6$ \\
        \rowcolor{blue!12}MambaAct L12 & $98.0$ & $99.8$ & $98.0$ & $94.0$ & $97.5$ & $10.81$ & $41.15$ & $56.06$ & $102.91$ \\
        \bottomrule
    \end{tabular}%
    }
    \vspace{-3mm}
\end{table}

\begin{table}[!t]
    \centering
    \caption{MambaAct composed with VLM-depth reduction on real-world
    tasks. Latency is measured on an RTX~5090 for the tube task; nut pouring
    shows similar latency.}
    \label{tab:ablation_real_world}

    \scriptsize
    \setlength{\tabcolsep}{2.2pt}
    \renewcommand{\arraystretch}{1.08}

    \resizebox{0.65\columnwidth}{!}{%
    \begin{tabular}{@{}lrrrrrr@{}}
        \toprule
        \multicolumn{7}{c}{\textbf{Model efficiency and latency}} \\
        \midrule
        Model &
        \shortstack{Params\\(B)} &
        \shortstack{Train\\(h)} &
        \shortstack{GFLOPs/\\query} &
        \multicolumn{3}{c}{Latency (ms)} \\
        \cmidrule(lr){5-7}
        & & & & Prefix & Denoise & E2E \\
        \midrule
        MambaAct
        & $2.96$ & $14.00$ & $4488.84$
        & $41.15$ & $31.11$ & $83.26$ \\

        \rowcolor{blue!12}
        MambaAct L12
        & $2.30$ & $13.55$ & $3209.70$
        & $32.05$ & $31.03$ & $74.19$ \\
        \bottomrule
    \end{tabular}%
    }

    \vspace{1mm}

    \resizebox{0.5\columnwidth}{!}{%
    \begin{tabular}{@{}llrrr@{}}
        \toprule
        \multicolumn{5}{c}{\textbf{Real-world success rates (\%)}} \\
        \midrule
        Task & Model & ID & OOD & Final \\
        \midrule
        \multirow{2}{*}{Nut pouring}
        & MambaAct
        & $100$ & $95.20$ & $96.90$ \\
        & \cellcolor{blue!12}MambaAct L12
        & \cellcolor{blue!12}$100$
        & \cellcolor{blue!12}$92.31$
        & \cellcolor{blue!12}$95.00$ \\
        \midrule
        \multirow{2}{*}{Tube placing}
        & MambaAct
        & $100$ & $82.70$ & $88.80$ \\
        & \cellcolor{blue!12}MambaAct L12
        & \cellcolor{blue!12}$100$
        & \cellcolor{blue!12}$84.62$
        & \cellcolor{blue!12}$90.00$ \\
        \bottomrule
    \end{tabular}%
    }

    \vspace{-3mm}
\end{table}
The same overall trend appears in real-world evaluation. MambaAct L12 reduces the deployed model size from $2.96$B to $2.30$B parameters and lowers RTX5090 inference latency from $83$ ms to $74$ ms. Despite this additional compression, it preserves $100\%$ ID success on both tasks and maintains comparable OOD performance, reaching $95.0\%$ final success on nut pouring and $90.0\%$ on tube placing.

\textbf{Ablation 3: Transferability.}
We test whether the Mamba action expert transfers beyond $\pi_{0.5}$ by replacing SmolVLA's Transformer action expert with a $4$-block Mamba expert. This reduces the action-expert size from $98.25$M to $64.31$M parameters, a $34.5\%$ reduction.

\begin{table}[!t]
    \centering
    \scriptsize
    \caption{Mamba action-expert transfer to SmolVLA on LIBERO. Both models are evaluated with $50$ trials per task and evaluation seed $7$ ($2{,}000$ episodes total), while retaining SmolVLA's native action horizon $H=50$, execution horizon $10$, and $10$ flow-matching denoising steps. Latency is measured on a single H100.}
    \label{tab:smolvla-mamba-ablation}
    \setlength{\tabcolsep}{1.6pt}
    \renewcommand{\arraystretch}{1.12}
    \resizebox{\columnwidth}{!}{%
    \begin{tabular}{@{}lcccccccccc@{}}
        \toprule
        & \multicolumn{5}{c}{LIBERO success rate (\%)} 
        & \multicolumn{3}{c}{Latency (ms)}
        & \multicolumn{1}{c}{Params} \\
        \cmidrule(lr){2-6}
        \cmidrule(lr){7-9}
        \cmidrule(lr){10-10}
        Model 
        & Spatial & Object & Goal & L-10 & Avg.
        & Prefix & Denoise & E2E
        & \shortstack{Action expert\\(M)} \\
        \midrule
        
        Released SmolVLA
        & $76.6$ & $93.0$ & $85.4$ & $70.2$ & $81.30$
        & $25.53$ & $163.15$ & $188.69$
        & $98.25$ \\
        
        \rowcolor{blue!12}
        SmolVLA-MambaAct
        & $82.6$ & $92.4$ & $87.0$ & $77.8$ & $\mathbf{84.95}$
        & $30.69$ & $59.89$ & $\mathbf{90.61}$
        & $64.31$ \\
        
        \bottomrule
    \end{tabular}%
    }
    \vspace{-6mm}
\end{table}

% Table~\ref{tab:smolvla-mamba-ablation} shows that MambaAct transfers effectively to SmolVLA. The published SmolVLA score of $87.3\%$ in Table~\ref{tab:libero-comparison} corresponds to the paper's $100$k-step simulation training setting with $10$ evaluation trials per task, whereas the publicly released LIBERO checkpoint was trained for $25$k steps with batch size $32$. We therefore re-evaluate this released $25$k checkpoint using our transfer protocol of $50$ trials per task with evaluation seed $7$ ($2{,}000$ episodes total), obtaining $81.30\%$ average LIBERO success. Training SmolVLA-MambaAct with the same documented $25$k-step, batch-$32$, seed-$1000$ training budget improves average success to $84.95\%$, a $3.65$ percentage-point gain. The largest improvements occur on LIBERO-10 ($70.2\%\rightarrow77.8\%$) and SPATIAL ($76.6\%\rightarrow82.6\%$), while OBJECT remains comparable. Because the published $87.3\%$ result and our $81.30\%$ re-evaluation use different training checkpoints and evaluation protocols, they are not directly comparable.
Table~\ref{tab:smolvla-mamba-ablation} shows that MambaAct transfers effectively to SmolVLA. The published SmolVLA score of $87.3\%$ in Table~\ref{tab:libero-comparison} uses the paper's $100$k-step training setting, whereas the released LIBERO checkpoint was trained for $25$k steps with batch size $32$. For a fair comparison, we therefore re-evaluate the released checkpoint under our protocol of $50$ trials per task with evaluation seed $7$, obtaining $81.30\%$ average success. Under the same $25$k-step, batch-$32$, seed-$1000$ training budget, SmolVLA-MambaAct reaches $84.95\%$ ($+3.65$ percentage points), while reducing denoising latency by $63.3\%$ and E2E latency by $52.0\%$. The largest gains occur on LIBERO-10 ($70.2\%\rightarrow77.8\%$) and SPATIAL ($76.6\%\rightarrow82.6\%$). 

These results indicate that the Mamba action-expert replacement transfers beyond the $\pi$-family, reducing action-expert size while improving average task success on a second flow-matching VLA backbone.

In summary, the ablations support three observations. First, Mamba substantially outperforms size-matched Transformer mixers under both evaluated conditioning interfaces, suggesting that the action mixer itself is important for preserving performance under aggressive expert compression. Second, gated conditioning provides an efficient interface between the pretrained VLM and the newly initialized action expert, although its accuracy advantage over cross-attention is partially confounded by model capacity. Third, the resulting Mamba action expert remains effective under VLM-depth reduction and transfers to a second flow-matching VLA architecture. Together, these results are consistent with benefits from selective action
sequence modeling and lightweight visual-language conditioning beyond
compression alone.
\FloatBarrier
\section{Conclusion}
We present and publicly release MambaAct, a lightweight Mamba-based action expert tailored for flow-matching vision-language-action models. Rather than altering the pretrained vision-language backbone or truncating denoising steps, MambaAct directly optimizes the frequently evaluated action-generation pathway. By substituting the Transformer action expert in $\pi_{0.5}$ with a Mamba mixer, MambaAct enables aggressive compression of the repeatedly evaluated action pathway, reducing the core action-expert parameter count from $311.5$M to $29.7$M ($\sim10\times$) while substantially lowering denoising latency. 
% Crucially, this compression comes with improved average performance on LIBERO alongside robust generalization across LIBERO-Plus and real-world UR5 manipulation tasks.
Crucially, this compression improves mean LIBERO performance across three
independent fine-tuning seeds, while retaining strong robustness on
LIBERO-Plus and on real-world robotic tasks. Furthermore, MambaAct seamlessly integrates with VLM-depth reduction and generalizes effectively when transferred to SmolVLA.

% Despite these strengths, several limitations remain. First, our real-world experiments are currently restricted to two UR5 tasks under primary spatial distribution shifts. In addition, while MambaAct composes well with VLM-depth reduction, its synergy with complementary efficiency techniques, such as visual-token pruning or step-compression mechanisms, remains to be empirically validated. Finally, while our findings confirm the empirical efficacy of the Mamba action expert, the precise contribution of individual state-space mechanisms within Mamba3 to these gains requires deeper isolation.

% Future work will focus on integrating MambaAct with prefix-side token pruning and denoising-step compression. 

 Despite these strengths, several limitations remain. First, our real-world experiments are currently restricted to two UR5 tasks under primary spatial distribution shifts. In addition, while MambaAct composes with VLM-depth reduction, its compatibility with other efficiency techniques, such as visual-token pruning and denoising-step compression, remains to be evaluated. We also leave deeper analysis of the gated visual-language conditioning and selective Mamba state dynamics to future work.

\bibliographystyle{IEEEtran}
\bibliography{references}
%%%%%%%%%%%%%%%%%%%%%%%%%%%%%%%%%%%%%%%%%%%%%%%%%%%%%%%%%%%%%%%%%%%%%%%%%%%%%%%%
\end{document}